\documentclass[sigconf]{acmart}
\usepackage{epsfig}
\usepackage{graphicx}
\usepackage{amsmath}
\usepackage{algpseudocode}
\usepackage{bm}

\usepackage{booktabs}
\usepackage{makecell}
\usepackage{multirow}
\usepackage{caption}
\usepackage{gensymb}
\usepackage[lined,boxed,commentsnumbered,ruled]{algorithm2e}
\usepackage{rotating}
\usepackage[caption=false]{subfig}
\usepackage{bigstrut} 
\usepackage{url}
\usepackage{balance}
\AtBeginDocument{%
  \providecommand\BibTeX{{%
    \normalfont B\kern-0.5em{\scshape i\kern-0.25em b}\kern-0.8em\TeX}}}

\copyrightyear{2022}
\acmYear{2022}
\setcopyright{acmcopyright}
\acmConference[MM '22] {Proceedings of the 30th ACM International Conference on Multimedia }{October 10--14, 2022}{Lisboa, Portugal.}
\acmBooktitle{Proceedings of the 30th ACM International Conference on Multimedia (MM '22), October 10--14, 2022, Lisbon, Portugal}
\acmPrice{15.00}
\acmISBN{978-1-4503-9203-7/22/10}
\acmDOI{10.1145/3503161.3547769}

\begin{document}
\title{Adaptive Mixture of Experts Learning for Generalizable Face Anti-Spoofing}



\author{Qianyu Zhou}
\authornote{Equal contributions. }
\authornote{Work done during an internship at Youtu Lab, Tencent.}
\orcid{0000-0002-5331-050X}
\affiliation{%
  \institution{Shanghai Jiao Tong University}
  \city{Shanghai}
  \country{China}
  \postcode{200240}
}
\email{zhouqianyu@sjtu.edu.cn}

\author{Ke-Yue Zhang}
\authornotemark[1]
\orcid{0000-0003-3589-5580}
\affiliation{
  \institution{Youtu Lab, Tencent}
  \city{Shanghai}
  \country{China}
  \postcode{200233}
}
\email{zkyezhang@tencent.com}

\author{Taiping Yao}
\authornotemark[1]
\orcid{0000-0002-2359-1523}
\affiliation{
  \institution{Youtu Lab, Tencent}
  \city{Shanghai}
  \country{China}
  \postcode{200233}
}
\email{taipingyao@tencent.com}

\author{Ran Yi}
\orcid{0000-0003-1858-3358}
\affiliation{%
  \institution{Shanghai Jiao Tong University}
  \city{Shanghai}
  \country{China}
  \postcode{200240}
}
\email{ranyi@sjtu.edu.cn}

\author{Shouhong Ding}
\affiliation{
  \institution{Youtu Lab, Tencent}
  \city{Shanghai}
  \country{China}
  \postcode{200233}
}
\email{ericshding@tencent.com}
\authornote{Corresponding Authors.}

\author{Lizhuang Ma}
\affiliation{%
  \institution{Shanghai Jiao Tong University}
  \city{Shanghai}
  \country{China}
  \postcode{200240}
}
\email{ma-lz@cs.sjtu.edu.cn}
\authornotemark[3]


\renewcommand{\shortauthors}{Qianyu Zhou, et al.}

\begin{abstract}
With various face presentation attacks emerging continually, face anti-spoofing (FAS) approaches based on domain generalization (DG) have drawn growing attention. Existing DG-based FAS approaches always
capture the domain-invariant features for generalizing on the various unseen domains. However, they neglect individual source domains’ discriminative characteristics and diverse domain-specific information of the unseen domains, and the trained model is not sufficient to be adapted to various unseen domains. To address this issue, we propose an Adaptive Mixture of Experts Learning (AMEL) framework, which exploits the domain-specific information to adaptively establish the link among the seen source domains and unseen target domains to further improve the generalization. Concretely,  Domain-Specific Experts (DSE) are designed to investigate discriminative and unique domain-specific features as a complement to common domain-invariant features. Moreover, Dynamic Expert Aggregation (DEA) is proposed to adaptively aggregate the complementary information of each source expert based on the domain relevance to the unseen target domain. And combined with meta-learning, these modules work collaboratively to adaptively aggregate meaningful domain-specific information for the various unseen target domains.  Extensive experiments and visualizations demonstrate the effectiveness of our method against the state-of-the-art competitors.
\end{abstract}




\maketitle

\section{Introduction}
\begin{figure}[t!]
	\centering
	\includegraphics[width=1.0\linewidth]{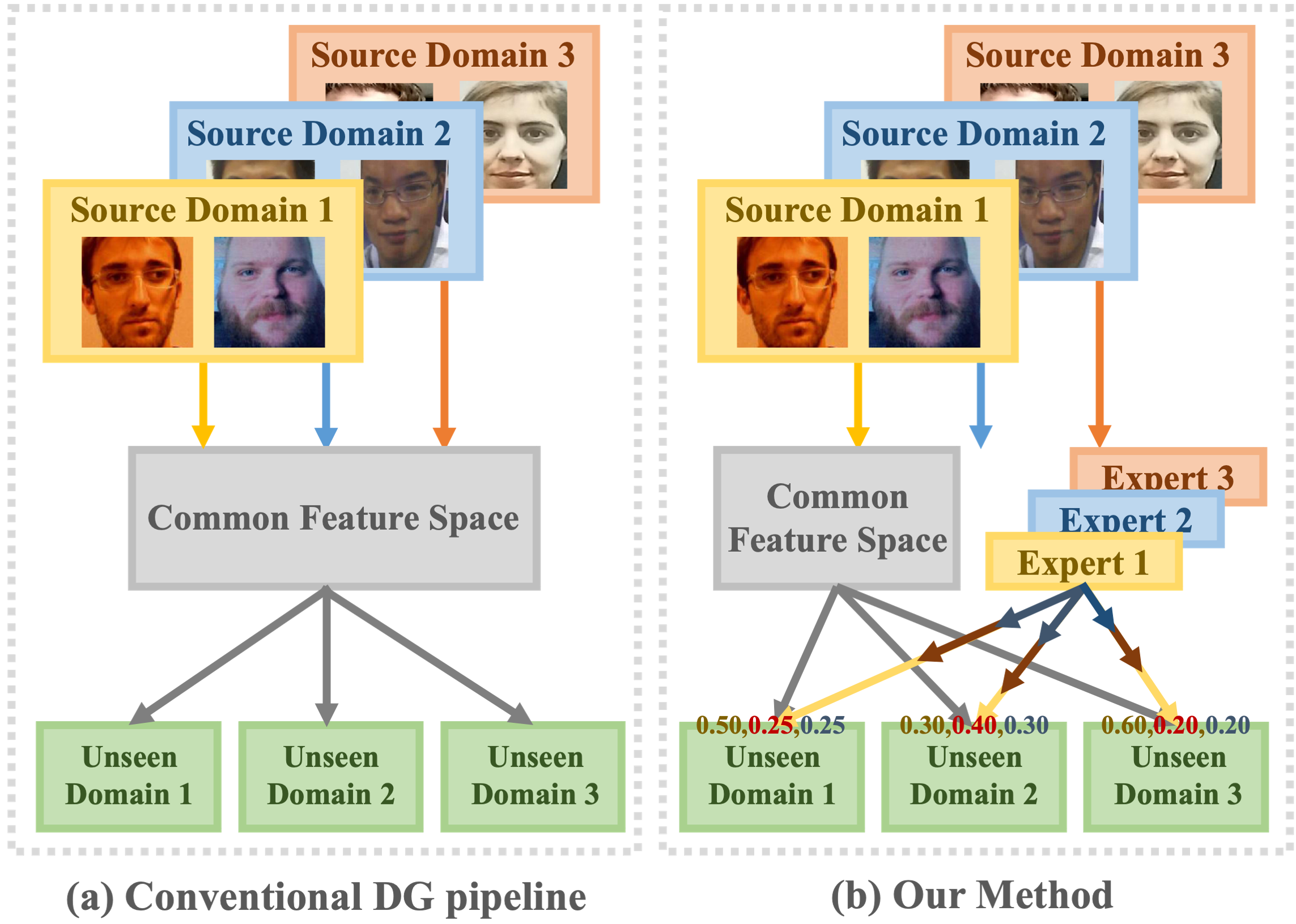}
	\caption{\textbf{Previous DG-based FAS methods map the images from multiple domains into a common feature space for learning domain-invariant features, while without considering the diverse domain-specific information of the unseen domains, the learned model is not sufficient to be adapted to various unseen domains. Instead, our method exploits the domain-specific information to adaptively establish the link between the seen source domains and unseen target domains to further improve the generalization.
} 
}
	\label{illustration}
\end{figure}
Face recognition (FR) techniques~\cite{deng2019arcface,kemelmacher2016megaface,taigman2014deepface,zhu2022local} have been widely utilized in identity authentication products, \emph{e.g.,} smartphones login, access control, entrance guard systems, \emph{etc}. 
Although FR systems bring great convenience, they are vulnerable to various face presentation attacks (PA), \emph{e.g.,} printed photo, video replay, and 3D masks. 
As such, face anti-spoofing (FAS) has been 
actively studied in recent years to detect these face presentation attacks, \emph{e.g.,} using hand-crafted features~\cite{2017Face,maatta2011face,LBP01,2014Context,HoG01,2016Secure}, deep-learning based features with binary-classification supervision~\cite{DeepBinary00,DeepBinary01,DeepBinary02,hashemifard2021compact}  or other auxiliary supervision~\cite{yu2021revisiting,zhang2021structure,lin2019face,CDCN,BCN}.
Although these methods achieve promising performance under intra-dataset scenarios, 
they often suffer from performance degradation when adapting to unseen domains due to the domain gap across different domains.

To improve the generalization,
recent studies~\cite{2019Multi,2020Single,2020Regularized,liu2021dual,liu2021adaptive,chen2021_D2AM} introduce the domain generalization (DG) techniques into FAS tasks. Based on the adversarial learning or meta-learning strategies, these DG-based FAS approaches endeavor to learn the domain-invariant features in a common feature space for generalizing to unseen target domains, as shown in Figure~\ref{illustration} (a). 
However, existing DG-based FAS methods suffer from two weaknesses: 
1) without considering the diverse domain-specific information of the target domains, the trained models are not sufficient to be adapted to various target domains; 
2) they often neglect individual source domains’ discriminative characteristics, which provides complementary and meaningful information especially when the unseen target domain is closely related to the source domains.

To address these issues, we propose a novel perspective of generalizable face anti-spoofing that exploits the domain-specific information to adaptively establish the link between the seen source and unseen target domains to further improve the generalization, as shown in Figure~\ref{illustration} (b).
Inspired by the works on mixture of experts (MoE) ~\cite{zhou2021domain,guo2018multi}, we propose Adaptive Mixture of Experts Learning (AMEL), a novel framework for generalizable face anti-spoofing. 
Specifically, we first learn the common domain-invariant features using a shared backbone with an instance normalization layer to mitigate the impact of domain information. 
Then, Domain-Specific Experts (DSE) are designed to investigate discriminative and unique domain-specific features as a complement to common domain-invariant features. 
Moreover, in order to establish the link with unseen target domains, Dynamic Expert Aggregation (DEA) is further proposed to adaptively aggregate the complementary information of each source expert based on the domain relevance \emph{w.r.t} the unseen target domain.
Furthermore, to simulate diverse target domains, we introduce the meta-learning strategy and a feature consistency loss to facilitate DSE to learn complementary meaningful information for the target domain and DEA to generate suitable aggregated weights. DSE contains several small residual blocks with a few parameters to ensure high efficiency.
As such, the proposed DSE and DEA work collaboratively to adaptively aggregate the specific information of the corresponding useful source domains for various unseen target domains, tackling the domain generalization issue in FAS. And to the best of our knowledge, this is the first work that treats generalizable FAS as a novel adaptive mixture of domain-specific experts paradigm. 

Our contribution  are summarized as follows:

$\bullet$ From a new perspective, 
we propose an Adaptive Mixture of Experts Learning (AMEL) framework for generalizable FAS,
which exploits the domain-specific information to adaptively establish the link between the seen source domains and unseen target domains to further improve the generalization.

$\bullet$ Combined with the meta-learning strategy and the feature consistency loss, Domain-Specific Experts (DSE) and Dynamic Expert Aggregation (DEA) are specifically designed to adaptively integrate source domain experts’ discriminative and meaningful information for various unseen target domains.
    
$\bullet$  
Extensive experiments and visualizations are presented to
reveal the role of  domain-specific information utilization, which demonstrates the
effectiveness of our method against state-of-the-art competitors on the widely-used benchmark datasets.

\begin{figure*}[t!]
	\centering
	\includegraphics[width=1.0\linewidth]{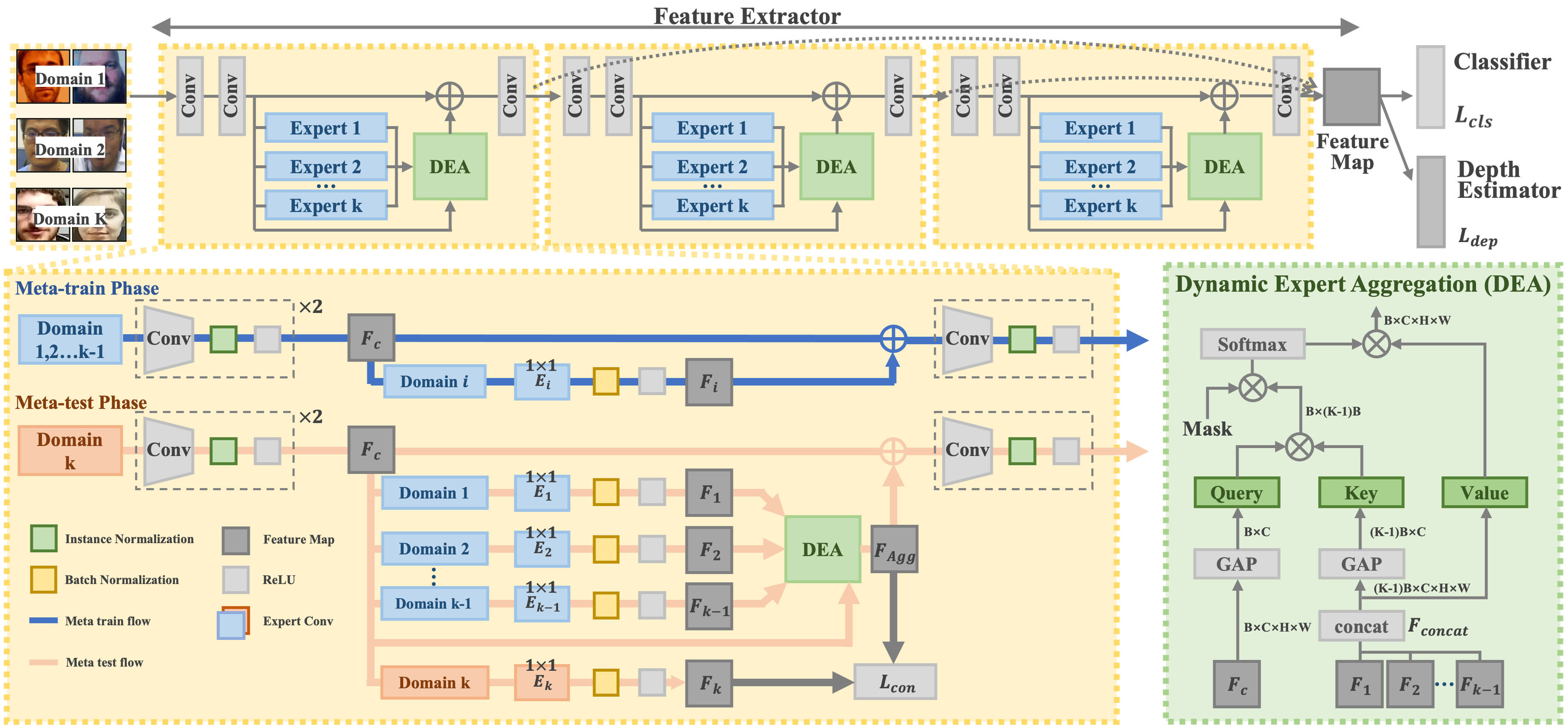}
	\caption{The pipeline of our proposed Adaptive Mixture of Experts Learning (AMEL) framework, which consists of three key components.
	First, our framework contains a common backbone with a instance normalization layer to mitigate the impact of domain information and extract the domain-invariant features.
	Then, based on the common features,  Domain-Specific Experts (DSE) contains several residual blocks $E_i$  
	to capture the individual domain's discriminative features $F_i$. 
	Moreover, Dynamic Expert Aggregation (DEA) dynamically leverage those complementary information of different domains into an aggregated feature based on the domain relevance \emph{w.r.t} the unseen domain.
	The meta-learning strategy combined with the feature consistency loss is adopted to simulate diverse unseen domains and further improve the generalization.
}
\label{fig:framework}
\end{figure*}

\section{Related Work}
\subsection{Face Anti-Spoofing}
Face anti-spoofing (FAS) has been actively studied in recent years to detect face presentation attacks.
Initially, researchers utilize handcrafted feature descriptors to extract information from images, such as LBP~\cite{LBP00,LBP01,maatta2011face} and SIFT~\cite{2016Secure}.
Considering that the facial motion may provide more discriminative information, some methods use the eyes blinking or the mouth speaking~\cite{siddiqui2016face,tu2019enhance} to 
detect the attacks.
Given the limited representation abilities of the handcrafted features, the results of the above methods are not satisfactory.
With the advent of deep learning, several approaches~\cite{DeepBinary00,DeepBinary01,DeepBinary02,2014Learn,zhang2021structure,wang2020deep} employ the convolutional neural network (CNN) to model the features, which gain larger improvements.
However, these methods always treat FAS as a binary classification task and lack sufficient supervision, easily leading the model to a local optimum.
Then, several methods~\cite{2018Learning,2018Face,additionInfo02,Yang_2019_CVPR, Zhang2021AuroraGR,chen2021dual} attempt to introduce different auxiliary supervisions to improve the results, such as depth map~\cite{yu2021revisiting}, reflection map~\cite{Kim2019BASNEF} and rppg signals~\cite{lin2019face,yu2019remote1,niu2020video,hu2021end}.
Based on such auxiliary supervisions, other approaches further explore specially-designed kernels~\cite{yu2021dual,yu2020multi} or the disentanglement framework~\cite{disentangle01,STCN} to extract more accurate features.
Although these methods achieve promising performance under intra-dataset scenarios, where the testing data comes from a similar distribution of the training data, they still suffer from performance degradation when adapting to unseen domains due to the domain gap across different domains. To improve the performance under the cross-dataset setting, several methods introduce unsupervised domain adaptation~\cite{DR-UDA,2018Unsupervised,2019Improving,jia2021unified,zhou2022generative} into the FAS area, while they rely on the unlabeled target domain during the training, which is always impractical in real-world scenarios.

\subsection{Generalizable Face Anti-Spoofing}
To adresss these issues, domain generalization~\cite{2019Multi,2020Single,2020Regularized,liu2021dual,liu2021adaptive,chen2021_D2AM} aims to learn a generalized model on the source domains without using the target domains, which is more practical and challenging for FAS.
Based on adversarial learning, 
MADDG~\cite{2019Multi} aligns all the samples from different domains to learn domain-invariant features.
While SSDG~\cite{2020Single} only maps the real samples of different domains to a common space without the fake ones.
Based on meta-learning, RFM~\cite{2020Regularized} aims to map the samples into a common classification space via a robust classifier and update with a more generalized optimization direction.  
D2AM~\cite{chen2021_D2AM}
learns the domain-invariant features better by
splitting the mixture source domains with maximum domain gaps.
Besides, DRDG~\cite{liu2021dual} and ANRL~\cite{liu2021adaptive} adaptively adjust the weights of samples and features or the coefficients of normalization to boost the alignment.
Although these methods gain a better generalization via 
learning the domain-invariant features, they neglect the diverse domain-specific information of the unseen target
domains and the trained model is still not sufficient to be
adapted to various unseen domains.
Besides, these methods overlook unique domain-specific
information of the source domains, which may provide complementary discriminative and meaningful information for generalizing to the target domains. In contrast, our proposed method exploits the domain-specific information to adaptively establish the link among the
seen source domains and unseen target domains to further improve the
generalization.

\subsection{Mixture of Experts}
Mixture of experts (MoE) is first introduced in ~\cite{jacobs1991adaptive}, which aims to learn a system composed of many separated networks (experts), where each expert learns to handle a subset of the whole dataset.
Equipped with deep learning, deep MoEs have shown their superiorities in image recognition~\cite{ahmed2016network,gross2017hard,wang2020deep}, image super-resolution~\cite{emad2022moesr}, image generation~\cite{shi2019variational}, scene parsing~\cite{fu2018moe}, machine translation~\cite{shazeer2017outrageously} and \emph{etc}.
Several methods~\cite{zhou2021domain,guo2018multi} utilize MoE to solve the domain adaptation (DA) via merging several domain experts to transfer toward the target domain.
Some works~\cite{dai2021generalizable, yan2021tal} also learn to aggregate the information from
multiple domain-specific networks for generalizable person re-identification. 
Inspired by them,  we propose a novel Adaptive
Mixture of Experts Learning (AMEL) framework, consists of specially designed  Domain-Specific
Experts (DSE) and Dynamic Expert Aggregation (DEA) for 
generalizable FAS.

\section{Methodology}
\subsection{Notations and Overview}
In the generalizable face anti-spoofing setting, we access to $K$ labeled source domains ${\bm{D}}=\{ {D}_{k} \}_{k=1}^{K} $, where $D_{k}=\{(x_{n}^{k},y_{n}^{k})\}_{n=1}^{N_{k}}$, $(x_{n}^{k},y_{n}^{k})$ is a labeled sample of domain $D_k$ and $N_{k}$ is the number of labeled images in the $k$-th domain. In this paper, $f$ denotes model functions and $F$ denotes the features.

Our proposed Adaptive Mixture of Experts Learning (AMEL) framework is shown in Figure~\ref{fig:framework}, which contains three key components: a shared feature extraction backbone, Domain-Specific Experts (DSE) and Dynamic Expert Aggregation  (DEA).
Specifically, the shared feature extraction backbone with a instance normalization layer is responsible to mitigate the impact of domain information and extract domain-invariant features.
Then, DSE is proposed to learn the discriminative and unique characteristics of each source domain as the complement to common domain-invariant features, and it contains several small residual blocks with a few parameters to ensure high efficiency. Moreover, to establish the
link with unseen target domains, DEA adaptively aggregates the complementary information of each source expert based on the domain
relevance with the unseen target domains. 
Furthermore, to simulate
diverse target domains, the meta-learning
strategy combined with a feature consistency loss is presented to facilitate the aggregation of
complementary meaningful information for the target domain to further improve the generalization.

\subsection{Domain-Specific Experts}
Most existing works~\cite{2019Multi,2020Single,2020Regularized,liu2021dual,liu2021adaptive,chen2021_D2AM} in DG FAS map the samples into a common feature space for capturing domain-invariant features to improve the generalization.
However, without considering the diverse domain-specific information of the target domains, the
trained models are not sufficient to be adapted to various target domains. Besides, the neglected domain-specific information of the source domains may be utilized to boost performances. In this work, the domain-specific features are the unique and complementary information of each source domain, such as domain-specific styles, and spoof-specific characteristics, which can further improve the generalization, especially when the unseen target domain is closely related to the source domains.

Specifically, we first utilize the instance normalization layer in the shared backbone to alleviate the impact of style information and learn the domain-invariant features.
Then, different from the above works, we further put forth Domain-Specific Experts (DSE) to extract discriminative and unique domain-specific features as the complement to domain-invariant features to facilitate the generalization.
Since training one individual model for each source domain to extract the domain-specific features requires large model parameters, the model size might become fairly large with the increase of source domains, limiting the practical deployment. 
Therefore, aiming to efficiently model the domain-specific features of each domain, here we implement DSE as $K$ residual blocks, where each residual block contains a small convolution layer with a kernel size of $1\times1$, a batch normalization layer and an activation layer (ReLU). 
Since DSE is much smaller than the common backbone, it has both low additional computational costs and little tendency to overfit. 
The whole process of each domain expert extracting the domain-invariant and domain-specific features are as follows:

\begin{equation}
\label{eq:residual}
f_k(x) =f_{\theta}(x)+ \Delta f_{\theta}^{k}(x),
\end{equation}
where $f_{\theta}$ represents the backbone that is shared by all source domains to learn the common domain-invariant features $F_{c}$ and each  $\Delta f_{\theta}^{k}$ is modeled as residual block to adatively extract the discriminative and unique domain-specific features $F_{k}$ of domain $D_{k}$.
To ensure the common backbone and DSE extract the task-related features, a binary classification loss $\mathcal{L}_{Cls}$ is defined as follows:
\begin{equation}
    \mathcal{L}_{Cls} = -\sum_{(x_n,y_n^{cls})} y_n^{cls} log(Cls(f_k(x_n))),
\end{equation}
where $Cls$ is the Binary Classifier detecting the face presentation attacks from the real ones, as shown in Figure~\ref{fig:framework}.
Since prior works~\cite{2018Learning,liu2021adaptive,liu2021dual} reveal that depth can be utilized as auxiliary information to supervise both live and spoof faces on pixel level, we strictly follow them using a Depth Estimator $Dep$ and estimate the facial depth maps for live faces and zero maps for spoof faces to facilitate the learning of Feature Extractor. Thus, $\mathcal{L}_{Dep}$ is formulated as follows:
\begin{equation}
    \mathcal{L}_{Dep} = \sum_{(x_n,y_n^{dep})} \left\| Dep(f_k(x_n)) - y_n^{dep} \right\|_{2}^{2},
\end{equation}
As shown in Figure~\ref{fig:framework}, the common backbone $f_{\theta}$ is updated via samples of all the source domains while each domain residual block $E_{i}$ in DSE is updated via the samples of corresponding domains $D_{i}$.

\subsection{Dynamic Expert Aggregation} 
Since different source domains have different domain relevance to the source domains, it is necessary to adaptively aggregate the information of each source expert based on the domain relevance. As such, we introduce a Dynamic Expert Aggregation (DEA) to adaptively establish the links among source domains and unseen target domains. Compared to the attention mechanism~\cite{dosovitskiy2020image,fan2021multiscale,liu2021swin} in Transformers that needs to interact token embeddings with each other, DEA instead queries the common features and only interacts with the feature of the domain experts, which largely simplifies the computation costs. In DEA, the inputs are the common feature $F_c$ and the concatenated features of experts $F_{concat}$, the output is the aggregated feature $F_{agg}$ of all these $K$ expert features during inference or $K-1$ expert features during meta-learning. Next, we elaborate on the aggregation of $K$ expert features.

Specifically, we perform several steps for feature aggregation. 
Firstly, we concatenate each domain-specific feature of different domain experts to obtain the features $F_{concat}$:
\begin{equation}
\label{eq:concat}
F_{concat}=concat(\Delta f_{\theta}^{1}(x), \Delta f_{\theta}^{2}(x), \cdots, 
\Delta f_{\theta}^{K}(x)),
\end{equation}
Then, the global average pooling of the common features $F_{c}$ and $F_{concat}$ is utilized as the query feature and key feature, respectively, and $F_{concat}$ is used as a value feature to compute the aggregated feature. 
These query, key and value features are forwarded to the DEA module as inputs:
\begin{equation}
\label{eq:dea}
F_{agg}=MaskedSoftmax({F_{c} \otimes F_{concat}^T}) \otimes  F_{concat},
\end{equation}
where MaskedSoftmax is the softmax function based on a binary mask $M$ with shape $(B, KB)$, $B$ is the batch size, $\otimes$ denotes the matrix multiplication. As \cite{dai2021generalizable} reveals that softmax is better than sigmoid in the voting mechanism, we use softmax in our method.
Concretely, the binary mask is generated by setting the relevant elements to value $1$, and all others to value $0$. 
\begin{equation}
M(i, j)=\left\{\begin{array}{l}
1, \text { if } j=kB+i, \text { For } k=0, 1, \cdots, K \\
0, \text { otherwise }
\end{array}\right.\end{equation}
where $i \in Range(0,B), j \in Range(0,KB)$. 
We iterate each location to generate the mask. 
Note that DEA is lightweight and does not require additional parameters during the training. 
Given the above steps, Eq.~\ref{eq:residual} is re-written as follows:
\begin{equation}
\label{eq:dynamic}
f_k(x)=f_{\theta}(x)+ \sum_{k=1}^{K} w^k \Delta f_{\theta}^{k}(x),
\end{equation}
where $w^k$ is the domain relevance between the $k$-th source domain and the unseen target domain. Note that we do not merely use domain-specific features to simulate target domains, but based on domain-invariant features, we use domain-specific features as additional complementary information that are beneficial for the discrimination in the target domain to improve the generalization.


\subsection{Optimization Strategies}
Due to the fact that meta-learning has shown its potential on enhancing the generalization abilities via the simulation of domain shifts among multi-source domains, we present a learning-to-learn strategy with a feature consistency loss to facilitate the whole framework to adaptively aggregate complementary meaningful domain-specific information for the various unseen target domains.
Here we elaborate the process of the optimization based on the meta-learning in detail, as shown in Algorithm~\ref{alg:algorithm1}.
At the beginning of each episodic training iteration, we randomly split the source domains into the meta-train set $\bm {D}_{trn}$ (simulated “source domains”) and the meta-test set $\bm {D}_{val}$ (simulated “unseen target domains”) to simulate the real-world scenarios.
Note that only domain-specific experts at residual blocks are updated via meta-learning strategy, while the other parameters in the base model including Feature Extractor, Binary Classifier and Depth Estimator follow the normal training process. 
Formally, we denote $\theta_{E}$ as the parameters of domain expert blocks and $\theta_{B}$ as the parameters of the base model, including the parameters of the shared Feature Extractor.

\noindent \textbf{Normal Train.}
Given batches sampled from all domains $D$, we calculate $\mathcal{L}_{B}$ to update $\theta_{B}$ for presentation attack detection:
\begin{equation}
\begin{split}
    &\mathcal{L}_{B}(\theta_{B},\theta_{E})= \sum_{D} \mathcal{L}_{Cls} + \mathcal{L}_{Dep} \\
    &\theta_{B} \leftarrow \theta_{B} - \beta \nabla_{\theta_{B}} \mathcal{L}_{B}(\theta_{B},\theta_{E})
\end{split}
\end{equation}

\noindent \textbf{Meta-Train.}  
During the meta-train phase, each residual block of the DSE is optimized separately to exploit the domain-specific information of the corresponding source domain via a classification loss $\mathcal{L}_{Cls}$ and depth estimation loss $\mathcal{L}_{Dep}$ similarly as the normal train phase on the meta-train set $D_{trn}$.

\begin{equation}
    \mathcal{L}_{trn}(\theta_{B},\theta_{E})= \sum_{D_{trn}} \mathcal{L}_{Cls} + \mathcal{L}_{Dep}.
\end{equation}
We optimize the learning direction of domain experts via calculating gradients of $\mathcal{L}_{trn}$, which is formulated as:
\begin{equation}
\begin{split}
    \theta_{E}' = \theta_{E} - \beta \nabla_{\theta_{E}}
    \mathcal{L}_{trn}(\theta_{B},\theta_{E})
\end{split}
\end{equation}
where $\theta_{E}^{'}$ denotes the updated parameters of $\theta_{E}$ after the one-step meta-training.

\noindent \textbf{Meta-Test.}
In the meta-test phase, the meta-test samples $x$ from domain $D_{k}$ will be forwarded to its own domain expert $E_{k}$ to get a extracted feature $F_{k}$, and meanwhile will be forwarded to other meta-train experts $\{ E_{i} \}_{i=1}^{K-1} \cup \{ E_{i} \}_{i=K+1}^{K}$ to generate a aggregated feature $F_{agg}$ with DEA module in Eq.~\ref{eq:dea}. 
Inspired by \cite{dai2021generalizable}, a feature consistency loss $\mathcal{L}_{con}$ is then imposed on these two latent features to push the aggregated feature $F_{agg}$ as discriminative as the extracted feature $E_{k}$, which forces the DEA modules to adaptively establish the link among the different source and target domains.
\begin{equation}
   \begin{aligned}
    & \mathcal{L}_{con} = \sum_{x_n \in D_{val}} \left\| E_j(x_n) - F_{agg}(x_n)) \right\|_{2}^{2} \\
    & \mathcal{L}_{val}(\theta_{B},\theta_{E}')= \sum_{D_{val}} \mathcal{L}_{Cls} + \mathcal{L}_{Dep}+  \lambda \mathcal{L}_{con},
    \end{aligned}
\end{equation}

\noindent \textbf{Meta-Optimization.}
In each iteration of episodic training, we obtain $\mathcal{L}_{trn}$ and $\mathcal{L}_{val}$ from meta-train and meta-test for optimization, which is formulated as below:
\begin{equation}
    \theta_{E} \leftarrow \theta_{E} - \gamma \nabla_{\theta_{E}} (\mathcal{L}_{trn}(\theta_{B},\theta_{E}) + \mathcal{L}_{val}(\theta_{B},\theta_{E}')).
\end{equation}
In the above training strategy, $\theta_{E}$ is updated via meta-learning and $\theta_{B}$ is optimized in the normal training process, which not only improves the generalization of our method, but also facilitates the stability and efficiency of meta-learning.
As such, the model obtains the abilities to aggregate the
complementary information for generalizable face anti-spoofing.

\begin{algorithm}[t!]
    \caption{Training Procedure of AMDE}
    \label{alg:algorithm1}
    \KwIn{
    Source domains $ \bm{D}=\{{D}_{i} \}_{i=1}^{K} $; Learning rate hyper-parameters $\beta, \gamma$; Balance hyper-parameter $\lambda$; MaxIters; MaxEpochs.
    }
    \KwOut{
    Backbone feature extractor; Domain-specific experts $\{E_{i}\}_{i=1}^{K}$; Classifier $Cls$; Depth Estimator $Dep$;
    }
    \For{$epoch=1$ to MaxEpochs}
    {
     \For{$iter=1$ to MaxIters}
        {
        Sample $K-1$ domains as meta-train set $\bm {D}_{trn}$ and the remaining domain as meta-test set $\bm {D}_{val}$;

        \textbf{Normal Train:} Sample batch in $D$ \;
        $\mathcal{L}_{base}(\theta_{B},\theta_{E})= \sum_{D} (\mathcal{L}_{Cls} + \mathcal{L}_{Dep}$) \;
        $\theta_{B} \leftarrow \theta_{B} - \beta \nabla_{\theta_{B}}
        \mathcal{L}_{base}(\theta_{B},\theta_{E})$ \;
        
        \textbf{Meta-Train:} Sample batch in meta-train set $D_{trn}$ \;
        
        $\mathcal{L}_{trn}(\theta_{B},\theta_{E}) = \sum_{D_{trn}} (\mathcal{L}_{Cls} + \mathcal{L}_{Dep})$ \;
        
        $\theta_{E}'=\theta_{E}-\beta \nabla_{\theta_{E}}\mathcal{L}_{trn}(\theta_{B},\theta_{E})$\;
        \textbf{Meta-Test:} 
        Sample batch in meta-test set $D_{val}$ \;
        
        $\mathcal{L}_{val}(\theta_{B},\theta_{E}') =  \sum_{D_{val}} (\mathcal{L}_{Cls}+\mathcal{L}_{Dep}+ 
        \lambda \mathcal{L}_{con}$)\;
        
        \textbf{Meta-Optimization:}
        
        $\theta_{E} \leftarrow \theta_{E} - \gamma \nabla_{\theta_{E}}(\mathcal{L}_{trn}(\theta_{B},\theta_{E})+
        \mathcal{L}_{val}(\theta_{B},\theta_{E}'))$ \;
        }
    }

\end{algorithm}

\begin{figure*}[t!]
\centering
\includegraphics[width=0.92\textwidth]{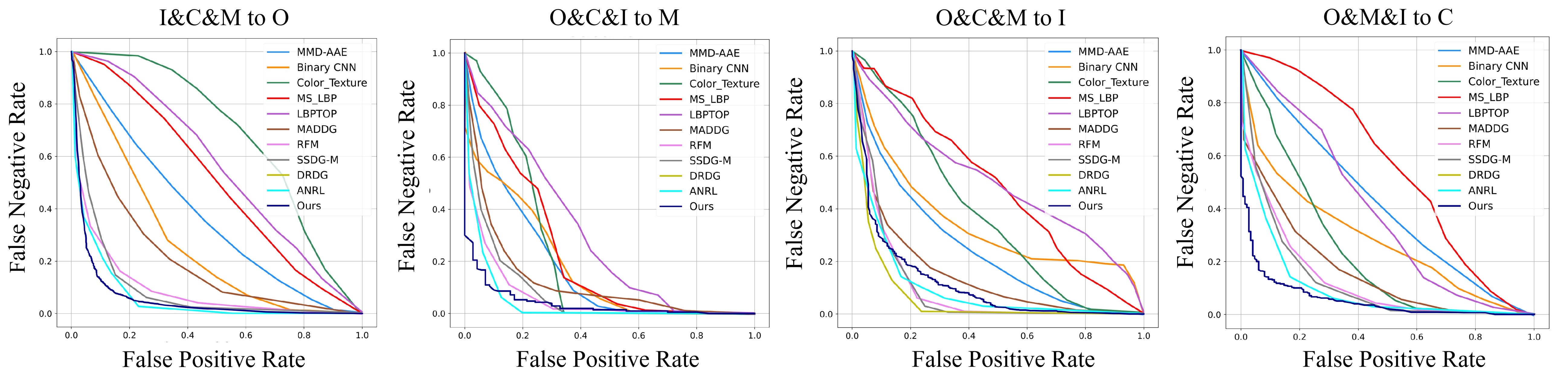}
\vspace{-3mm}
\caption{ROC curves compared to state-of-the-art FAS approaches.}
\label{fig:roc}
\end{figure*}

\begin{table*}[t!]
\centering
\begin{center}
\vspace{-1mm}
\caption{Comparison to the-state-of-art FAS methods on four testing domains. The bold type indicates the best performance.}
\vspace{-2mm}
\label{tab:DG_SOTA_3to1}
\resizebox{0.91\textwidth}{!}{%
\begin{tabular}{c | c c | c c | c c | c c }
\toprule
\multirow{2}{*}{\textbf{Methods}} &
\multicolumn{2}{c|}{\textbf{I\&C\&M to O}} &
\multicolumn{2}{c|}{\textbf{O\&C\&I to M}} &
\multicolumn{2}{c|}{\textbf{O\&C\&M to I}} &
\multicolumn{2}{c}{\textbf{O\&M\&I to C}} \\

&HTER(\%) &AUC(\%) &HTER(\%) &AUC(\%) &HTER(\%) &AUC(\%) &HTER(\%) &AUC(\%)\\
\midrule
IDA~\cite{2015Face}  &$54.20$ &$44.59$ &$66.67$ &$27.86$ &$28.35$ &$78.2$5 &$55.17$ &$39.05$ \\
LBPTOP~\cite{2014dynamic}&$53.15$ &$44.09$  &$36.90$ &$70.80$  &$49.45$ &$49.54$ &$42.60$ &$61.05$ \\
MS\_LBP~\cite{maatta2011face}&$50.29$ &$49.31$  &$29.76$ &$78.50$  &$50.30$ &$51.64$ &$54.28$ &$44.98$\\
ColorTexture~\cite{2017Face}&$63.59$ &$32.71$  &$28.09$ &$78.47$  &$40.40$ &$62.78$ &$30.58$ &$76.89$\\
Binary CNN~\cite{2014Learn}&$29.61$ &$77.54$  &$29.25$ &$82.87$  &$34.47$ &$65.88$ &$34.88$ &$71.94$\\
Auxiliary (ALL)~\cite{2018Learning}&- &- &- &-  &$27.60$ &- &$28.40$ &- \\
Auxiliary (Depth)~\cite{2018Learning}&$30.17$ &$77.61$  &$22.72$ &$85.88$  &$29.14$ &$71.69$ &$33.52$ &$73.15$\\
\midrule
MMD-AAE~\cite{2018Domain}&$40.98$ &$63.08$  &$27.08$ &$83.19$  &$31.58$ &$75.18$ &$44.59$ &$58.29$\\
MADDG~\cite{2019Multi}&$27.98$ &$80.02$  &$17.69$ &$88.06$  &$22.19$ &$84.99$ &24.50 &84.51\\
RFM~\cite{2020Regularized}&16.45 &91.16  &13.89 &93.98  &17.30 &90.48 &20.27 &88.16\\
SSDG-M~\cite{2020Single}& 25.17 & 81.83  &16.67 &90.47  & 18.21 &\textbf{94.61} & 23.11 & 85.45 \\
D$^2$AM~\cite{chen2021_D2AM} &15.27 &90.87  &12.70 &95.66 &\textbf{15.43 }&91.22 &20.98 &85.58\\
DRDG~\cite{liu2021dual}&15.63 & 91.75  & 12.43 & 95.81  & 15.56 & 91.79  & 19.05 & 88.79\\
ANRL~\cite{liu2021adaptive}& 15.67 & 91.90  & 10.83 & \textbf{96.75}  & 16.03 & 91.04  & 17.85 & 89.26\\

\midrule
Ours (AMEL) & \textbf{11.31} & \textbf{93.96}  &\textbf{10.23} &96.62 &18.60 &88.79 &\textbf{11.88} &\textbf{94.39}\\
\bottomrule
\end{tabular}}
\end{center}
\vspace{-1mm}
\end{table*}

\begin{table}[t!]
\centering
\begin{center}
\caption{Comparison results on limited source domains.}
\vspace{-1mm}
\label{tab:DG_SOTA_2to1}
\begin{tabular}{c | c c | c c }
\toprule
\multirow{2}{*}{\textbf{Methods}} &
\multicolumn{2}{c|}{\textbf{M\&I to C}} &
\multicolumn{2}{c}{\textbf{M\&I to O}} \\
    &HTER(\%) &AUC(\%) &HTER(\%) &AUC(\%) \\
\midrule
MS\_LBP~\cite{maatta2011face} & $51.16$ & $52.09$ &  $43.63$ & $58.07$ \\
IDA~\cite{2015Face} & $45.16$ & $58.80$ &  $54.52$ & $42.17$ \\
Color Texture~\cite{2017Face} & $55.17$ & $46.89$ &  $53.31$ & $45.16$ \\
LBPTOP~\cite{2014dynamic} & $45.27$ & $54.88$ &  $47.26$ & $50.21$ \\
MADDG~\cite{2019Multi} & $41.02$ & $64.33$ &  $39.35$ & $65.10$ \\
SSDG-M~\cite{2020Single} & $31.89$ & $71.29$ &  $36.01$ & $66.88$ \\
D$^2$AM~\cite{chen2021_D2AM} &32.65 &72.04 &27.70 & 75.36 \\
DRDG~\cite{liu2021dual} &31.28 &71.50 &33.35 & 69.14 \\
ANRL~\cite{liu2021adaptive} & 31.06 & 72.12 & 30.73 &74.10 \\
\midrule
Ours (AMEL) &\textbf{23.33} &\textbf{85.17} & \textbf{19.68} & \textbf{87.01}  \\
\bottomrule
\end{tabular}
\end{center}
\vspace{-4mm}
\end{table}

\section{Experiments}
In this section, we first describe the experimental setup in Section~\ref{sec:4.1}, including the benchmark datasets and the implementation details. 
Then, in Section~\ref{sec:4.2}, we demonstrate the effectiveness of our framework compared to the state-of-the-art approaches on six benchmark datasets. 
Next, in Section~\ref{sec:4.3}, we conduct ablation studies to investigate the role of each component in the method. 
Finally, we provide detailed visualization and analysis in Section~\ref{sec:4.4}.


\subsection{Experimental Setting}
\label{sec:4.1}
\noindent \textbf{Datasets.} 
We use four public datasets that are widely-used in FAS research to evaluate the effectiveness of our method: OULU-NPU \cite{2017OULU} (denoted as O), CASIA-MFSD \cite{Zhang2012A} (denoted as C), Idiap Replay-Attack \cite{2012Replay} (denoted as I), and MSU-MFSD \cite{2015Face} (denoted as M). 
These four datasets are collected with diverse capture devices, attack types, illumination conditions, background scenes, and races. 
Therefore, significant domain shifts exist among these datasets.
In all experiments, we strictly follow the same protocols as previous DG methods~\cite{2019Multi,2020Single,2020Regularized,liu2021dual,liu2021adaptive} in FAS for experiments.

\noindent \textbf{Implementation Details.} 
Our method is implemented via PyTorch~\cite{paszke2019pytorch} on $24G$ NVIDIA $3090$ GPUs and trained with Adam optimizer~\cite{kingma2014adam}. 
We use the same backbone MADDG (M) as existing works~\cite{liu2021dual,liu2021adaptive,2020Regularized}. 
We extract RGB channels of images, thus the input size is $256\times256\times3$. 
For training, the hyper-parameter $\lambda$ is set to $0.1$ respectively. 
Both learning rates $\beta$ are $\gamma$ are set to $0.0001$.
Following prior works~\cite{liu2021adaptive,liu2021dual}, we utilize the PRNet~\cite{2018Joint3D} to generate the pseudo-depth maps. 
The Half Total Error Rate (HTER) and the Area Under Curve (AUC) are used as the evaluation metric~\cite{2019Multi}.

\subsection{Comparison to the State-of-the-art Methods}
\label{sec:4.2} To validate the generalization capability towards the unseen target domain on the FAS task, we perform two experiments following common DG FAS protocols, \emph{i.e.,} Leave-One-Out (LOO) generalization and generalization with limited source domains, respectively. 

\noindent \textbf{Comparisons in Leave-One-Out (LOO) Setting.} As shown in Table~\ref{tab:DG_SOTA_3to1}, we conduct cross-domain generalization in four common Leave-One-Out (LOO) settings of FAS task. We randomly select three datasets
from them as source domains and the left one is treated as unseen target domain, which is unavailable during the training process.
The comparison methods in Table~\ref{tab:DG_SOTA_3to1} are divided into two parts: conventional FAS methods and DG-based FAS methods.
From the table, we make the following observations. (1). Conventional FAS methods~\cite{2015Face,2014dynamic,maatta2011face,2017Face,2014Learn,2018Learning} show less-desired performances than DG-based FAS methods~\cite{2018Domain,2019Multi,2020Single,2020Regularized,liu2021adaptive,liu2021dual} under these four cross-dataset benchmarks, this is due to they
do not consider learning the domain-invariant representations across domains. (2). Our method outperforms most of these DG-based methods under four test settings. The main reason is that all these methods endeavor to construct a shared feature space across different domains and they largely neglect the source domains' discriminative and complementary characteristics, while our proposed method successfully address these two issues.

\noindent \textbf{Comparisons on Limited Source Domains.}
As illustrated in Table~\ref{tab:DG_SOTA_2to1}, we also evaluate our method when extremely limited source domains (\textit{i.e.}, only two source datasets) are available. Following \cite{liu2021dual,liu2021adaptive}, MSU-MFSD (M) and ReplayAttack (I) datasets are selected as the source domains for training, and the remaining two ones, \textit{i.e.}, CASIA-MFSD (C) and OULU-NPU (O), respectively, are used as the target domains for testing. Our proposed method outperforms the state-of-the-art approaches by a large margin despite limited source data in this more challenging case, which powerfully verifies its efficiency and generalizability on unseen target domains. 

\begin{table}[t]
\caption{Ablation of each component on I$\&$C$\&$M to O.}
\label{table:ablation_component}
\centering
\begin{tabular}{cccc|c|c} \toprule
ID & Baseline & DSE+DEA & Meta & HTER(\%) & AUC(\%)\\
\midrule
I & \checkmark  & - & - & 19.72 & 87.58 \\
II & \checkmark   & \checkmark  & - & 14.74 & 91.90 \\
III & \checkmark  & \checkmark  &  \checkmark & 11.31 & 93.96 \\
\bottomrule
\end{tabular}
\vspace{-1mm}
\end{table}

\begin{figure}[t!]
\centering
\includegraphics[width=0.5\textwidth]{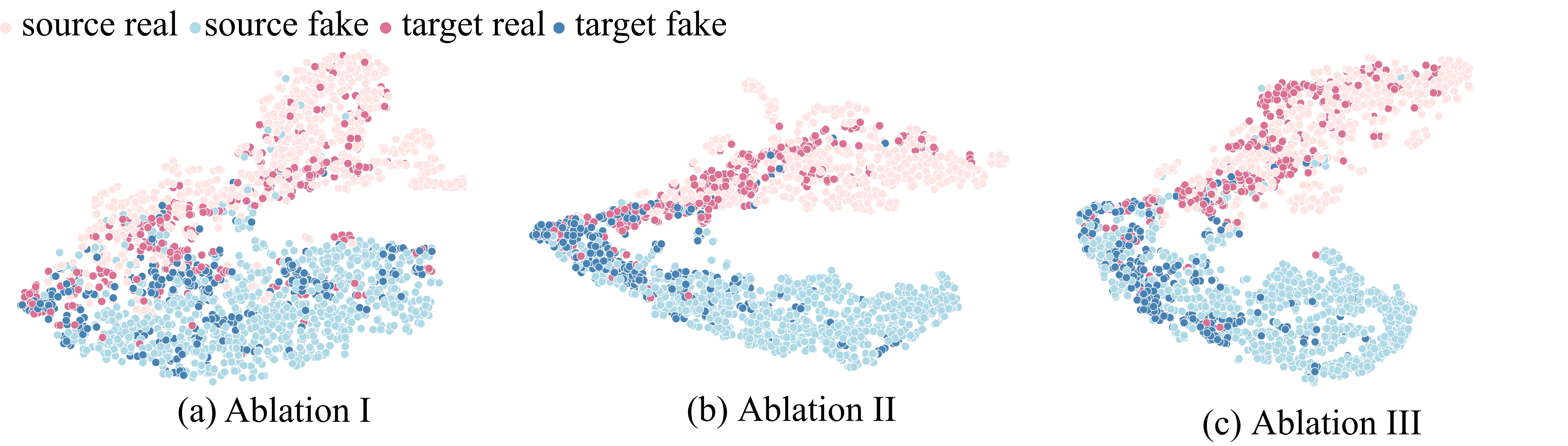}
\caption{The t-SNE feature visualization of ablation studies.}
\label{fig:tsne_ablation}
\vspace{-3mm}
\end{figure}

\subsection{Ablation Studies}
\label{sec:4.3} In this section, we perform ablation experiments to investigate the contribution of each component.
We also investigate the effect of different aggregation strategies, expert designs, and inference strategies to demonstrate the effectiveness. All the ablation experiments are conducted on  the I\&C\&M to O setting.

\noindent \textbf{Effectiveness of each component.} 
Table~\ref{table:ablation_component} shows the ablation studies of each component. The baseline means training the MADDG backbone~\cite{2019Multi,liu2021dual,liu2021adaptive} with IN layers, and the results are with $19.72\%$ HTER and $87.58\%$ AUC. 
By adding DSE and DEA, we boost the HTER and AUC performance by an additional $4.98\%$, $4.32\%$, achieving $14.74\%$ HTER and $91.90\%$ AUC, respectively.  Finally, our proposed meta-learning algorithm including the feature consistency loss can effectively lower the HETR to $11.31\%$ and increase the AUC to $93.96\%$, respectively.
These improvements validate the effectiveness of individual components of our proposed approach. It also reveals that these components are complementary and together they significantly promote performance.

As shown in Figure~\ref{fig:tsne_ablation}, we visualize the feature distributions to further demonstrate the effectiveness of each part in our proposed method.  From Figure~\ref{fig:tsne_ablation} (a), as for the baseline model, the source features are well-discriminated while the unseen target features are not. 
As shown in Figure~\ref{fig:tsne_ablation} (b), by adding DSE for each domain together with DEA, the above issue is alleviated to some extent. The classification boundary becomes more clear but there are still some samples misclassified near the decision boundary since the aggregation abilities are not fully generalizable due to the lack of simulating the unseen domains.
As such, by further adding meta-learning with the feature consistency loss in Figure~\ref{fig:tsne_ablation} (c), our approach manages to learn a better decision boundary between these two categories.

\begin{table}[t]
\caption{Ablation of expert designs on I$\&$C$\&$M to O.}
\label{table:ablation_expert_design}
\centering
\begin{tabular}{c|c|c} 
\toprule
Expert Designs & HTER(\%) & AUC(\%) \\
\midrule
IN+Conv+Relu &16.25 &90.90  \\
BN+Conv+Relu &17.07  &90.68  \\
Conv+IN+Relu &13.71 &92.97 \\
\midrule
Ours (Conv+BN+Relu) & 11.31 & 93.96  \\
\bottomrule
\end{tabular}
\vspace{-1mm}
\end{table}

\begin{table}[t]
\caption{Ablation of aggregation strategies on I$\&$C$\&$M to O.}
\label{table:ablation_aggregation}
\centering
\begin{tabular}{c|c|c} 
\toprule
Aggregation Strategy & HTER(\%) & AUC(\%)\\
\midrule
Average Voting &17.31 & 90.46 \\
Expert Ensembling &18.98  &89.50  \\
Max Selection &14.89  &92.42  \\
\midrule
Ours & 11.31 & 93.96 \\
\bottomrule
\end{tabular}
\vspace{-3mm}
\end{table}

\noindent \textbf{Ablations of different expert designs.}
\label{sec:expert ablation}
Table~\ref{table:ablation_expert_design} illustrates the comparisons of different expert designs. For a lightweight design, we implement our expert with a 1$\times$1 convolutional block, BN layer, and a relu activation. 
When using an IN+Conv+Relu or Conv+IN+Relu, the performance is less-desirable in terms of both HTER and AUC. The main reasons are that IN layer would filter some domain-specific information, \emph{e.g.,} styles, and it is not beneficial for explor the source domains' unique and discriminative information. Compared to them,  we find Conv+BN+Relu is the best combination. 

\begin{table*}[t!]
\centering
\begin{center}
\caption{Comparison of inference strategies on four testing domains. The bold type indicates the best performance.}
\label{table:ablation_inference}
\resizebox{0.97\textwidth}{!}{%
\begin{tabular}{c | c c | c c | c c | c c }
\toprule
\multirow{2}{*}{\textbf{Methods}} &
\multicolumn{2}{c|}{\textbf{I\&C\&M to O}} &

\multicolumn{2}{c|}{\textbf{O\&C\&I to M}} &
\multicolumn{2}{c|}{\textbf{O\&C\&M to I}} &
\multicolumn{2}{c}{\textbf{O\&M\&I to C}} \\

&HTER(\%) &AUC(\%) &HTER(\%) &AUC(\%) &HTER(\%) &AUC(\%) &HTER(\%) &AUC(\%)\\
\midrule
Common feature + Expert O &--     & --    &17.38 &90.89  &28.50 &79.04  &16.00  &92.45\\
Common feature + Expert M & 23.86 & 83.58 &--     &--    &\textbf{18.40} &87.14  &19.89  &88.64\\
Common feature + Expert I & 29.00 & 75.75 &22.14 &87.27 &-- &-- &23.56  &86.71 \\
Common feature + Expert C & 13.53 & 92.23  &15.00 &93.18 &26.10 &76.36 &--     &--\\
\midrule
Ours (Common feature + Aggregated feature) & \textbf{11.31} & \textbf{93.96}  &\textbf{10.23} &\textbf{96.62} &18.60 &\textbf{88.79} &\textbf{11.88} &\textbf{94.39} \\
\bottomrule

\end{tabular}}
\end{center}
\end{table*}

\noindent \textbf{Impact of aggregation strategies.} Table~\ref{table:ablation_aggregation} shows the impact of different aggregation strategies on the same baseline. Average Voting means to average the domain relevance of all domain experts to make final aggregation, and Expert Ensembling means to directly sum the domain relevance of all expert branches. Max Selection means to select the maximum value of the domain relevance, in other words, the most similar source domain expert with the unseen target domain will be utilized for feeding forward.
From the table, we observe that our proposed DEA outperforms these two related aggregation strategies. This is because their aggregation strategies largely overlook the domain relevance \emph{w.r.t} the unseen target domain, and naively averaging or ensembling all the source domains' individual information is impractical.

\begin{figure}[t!]
\centering
\includegraphics[width=0.45\textwidth]{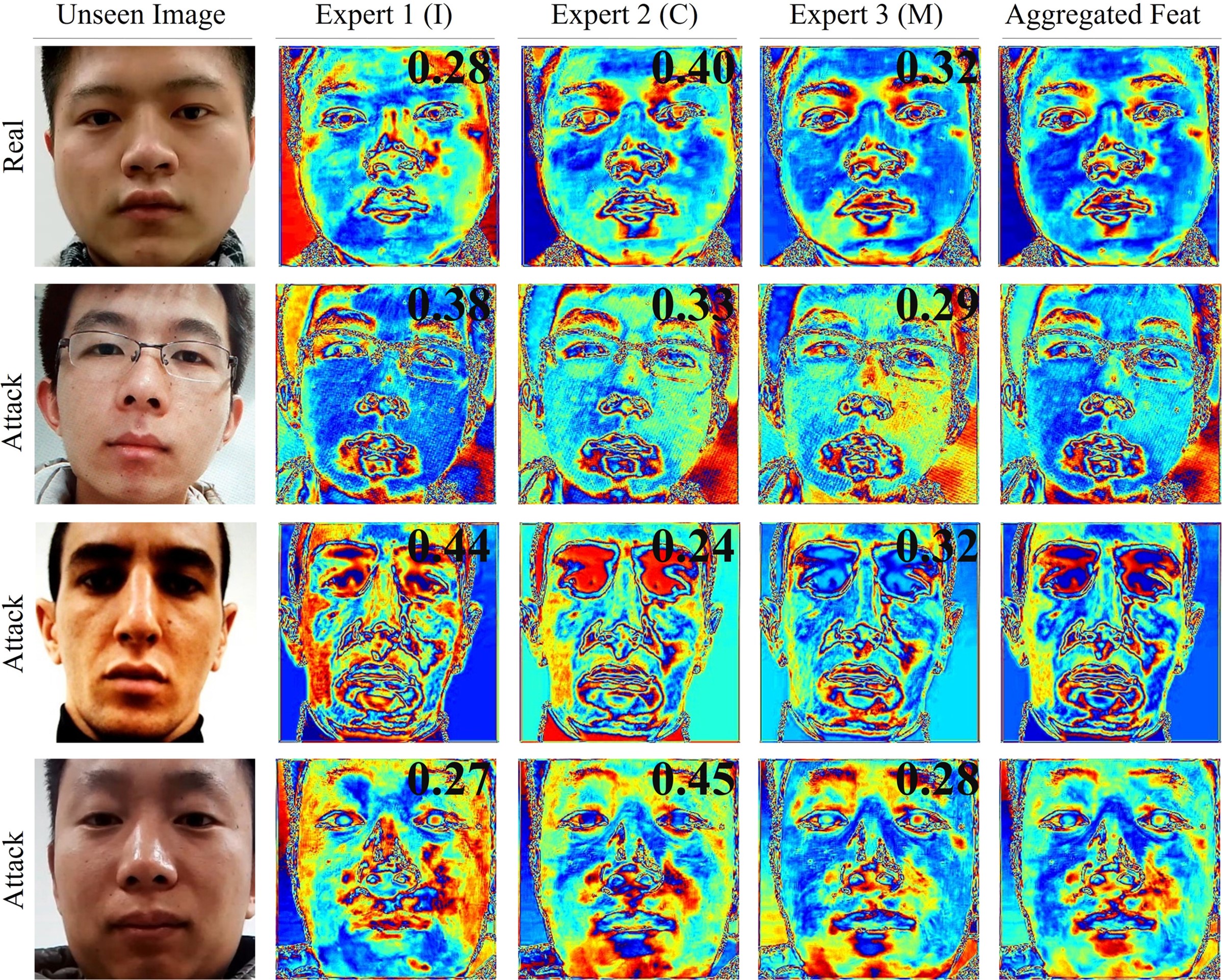}
\caption{Visualization on features of different domain experts and the aggregated feature on the I\&C\&M to O setting.}
\vspace{-3mm}
\label{fig:vis_res}
\end{figure}

\subsection{Visualization and Analysis}
\label{sec:4.4}

 \noindent \textbf{Analysis of different inference strategies using the corresponding expert branch.} As shown in Table \ref{table:ablation_inference}, when using the source domain expert branch with the common feature for inference, the performance are less-desired, which indicates that only using one single source expert is not sufficient for generalization due to the large domain shifts. Besides, we can find that the not all source domains contribute equally towards the unseen target domain. For example, in I\&C\&M to O setting, the expert branch of C dataset could achieve better performance than others on the unseen target domain due to the domain relevance between C dataset and O dataset are higher than others. Compared to these aforementioned inference strategies, our proposed approach can achieve the best performance on four challenging benchmark datasets.

\noindent \textbf{Visualizations of expert feature and aggregated features}. As shown in Figure~\ref{fig:vis_res}, we visualize the features of each source domain expert and the aggregated feature in the I\&C\&M to O setting. And we can make the following observations: (1). Each domain expert learns differently, and
those source samples that share more similarities with the unseen target samples will provide more meaningful and discriminative information for enhance the generalization. 
For example, in the third row, the expert of domain I and  domain C pay attention to the eye region, while the expert of domain M ignores the region.
(2). Since each sample in the domain O has different similarities with three different source domains, the weights of each samples are various, which further confirms that our method could dynamically utilize the specific source information to adaptively simulate the domain-specific features of the unseen target domain. 

\begin{figure}[t!]
\centering
\includegraphics[width=0.48\textwidth]{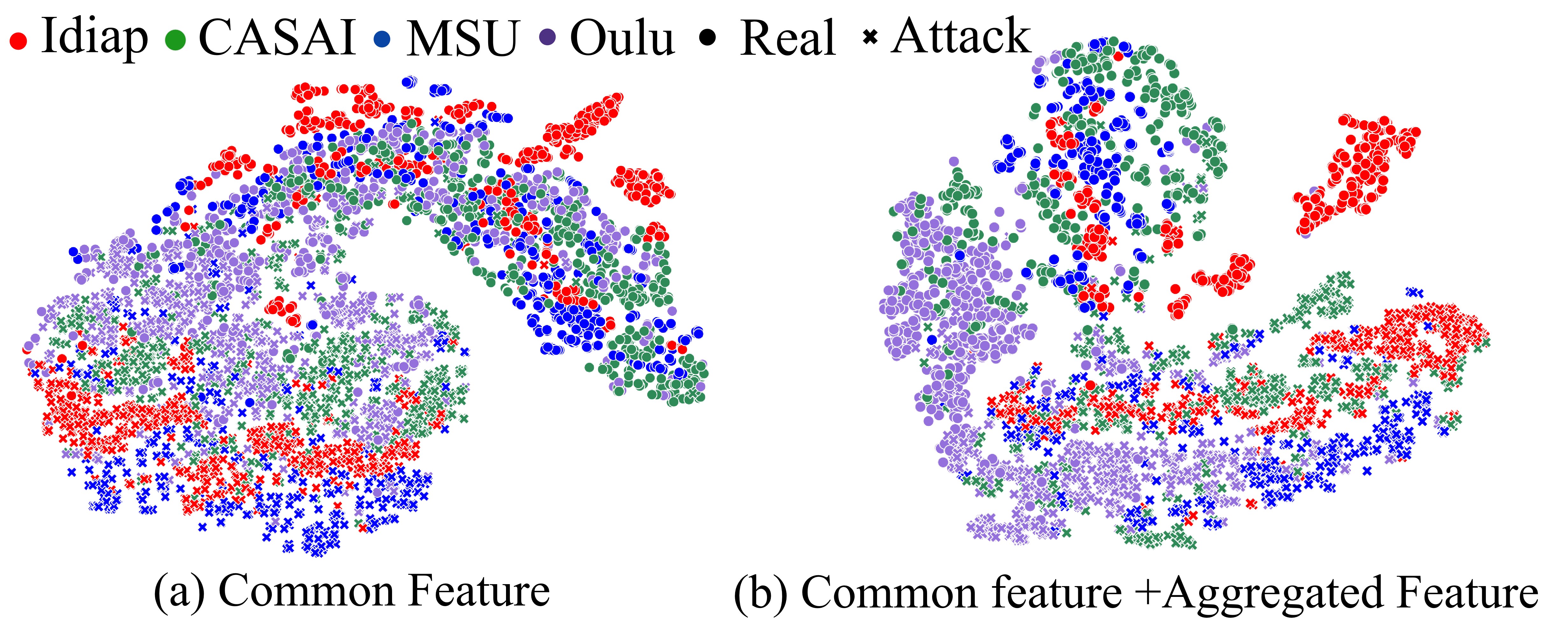}
\caption{The t-SNE visualization of common features and aggregated features on I\&C\&M to O setting.}
\vspace{-9mm}
\label{fig:tsne}
\end{figure}

\noindent \textbf{The t-SNE visualization of features.} 
To understand how the AMEL framework aligns the feature, we utilize
t-SNE to visualize the feature distributions of each domain, including three source domains and the unseen target domain in the I\&C\&M to O setting.
We can make the following two observations: (1). When using the common feature for generalization, as shown in Figure~\ref{fig:tsne} (a), the source data can be well discriminated by binary classification. 
However, the target data is not well-classified near the decision boundary. Instead, in Figure~\ref{fig:tsne}(b) our approach manages to learn a better decision boundary between these two categories of real and attack samples. (2). Besides, the common features of the same category are more compact, while the distributions of aggregated feature are looser. 
The main reason is that merely learning the domain-invariant feature in the common feature space is not sufficient for domain generalization, and those discriminative and unique domain-specific features could improve the performance as a useful complement to the common domain-invariant features. 

\section{Conclusion}
In this paper, we present a novel Adaptive Mixture of Experts Learning (AMEL) framework to address the generalizable face anti-Spoofing, which exploits the domain-specific information to adaptively establish the link between the seen source and unseen target domains to further improve the generalization.
Specifically,  Domain-Specific Experts(DSE) are trained to learn the individual domain's discriminative feature. 
Besides, Dynamic Expert Aggregation (DEA) is presented to dynamically leverage complementary information of different domains into an aggregated feature based on the domain relevance \emph{w.r.t} the unseen domain. 
Finally, to enhance the model generalizability, we present a meta-learning algorithm with a feature consistency loss to simulate the domain-specific feature of the unseen target domain. 
Extensive experiments with analysis demonstrate the effectiveness of our proposed approach.

\begin{acks}
This work is supported by National Key Research and Development Program of China (2019YFC1521104), National Natural Science Foundation of China (72192821, 61972157), Shanghai Municipal Science and Technology Major Project  (2021SHZDZX0102), Shanghai Science and Technology Commission (21511101200, 22YF1420300), and Art major project of National Social Science Fund (I8ZD22). 
\end{acks}


\bibliographystyle{ACM-Reference-Format}
\balance
\bibliography{sample-base}










\end{document}